\documentclass[10pt,twocolumn,letterpaper]{article}

\usepackage{cvpr}
\usepackage{times}
\usepackage{epsfig}
\usepackage{graphicx}
\usepackage{amsmath}
\usepackage{amssymb}
\usepackage{subfig}
\usepackage{multirow}
\usepackage{algorithm}
\usepackage{algorithmic}
\usepackage[titletoc]{appendix}

\usepackage[breaklinks=true,bookmarks=false]{hyperref}

\cvprfinalcopy 

\def\cvprPaperID{****} 

\begin{document}

\title{Face-NMS: A Core-set Selection Approach for Efficient Face Recognition}

\author{Yunze Chen\textsuperscript{\rm 12}\thanks{These authors contributed equally.},
    Junjie Huang\textsuperscript{\rm 1}\footnotemark[1] \thanks{Corresponding author.},
    Jiagang Zhu\textsuperscript{\rm 1},
    Zheng Zhu\textsuperscript{\rm 3},
    Tian Yang\textsuperscript{\rm 1},
    Guan Huang\textsuperscript{\rm 1},
    Dalong Du\textsuperscript{\rm 1}\\
    \textsuperscript{\rm 1} XForwardAI Technology Co.,Ltd, Beijing, China\\
    \textsuperscript{\rm 2} Institute of Automation, Chinese Academy of Sciences, Beijing, China\\
    \textsuperscript{\rm 3} Tsinghua University, Beijing, China\\
{\tt\small chenyunze2018@ia.ac.cn, \{junjie.huang, zhengzhu\}@ieee.org,} \\ {\tt\small\{jiagang.zhu, tian.yang, guan.huang, dalong.du\}@xforwardai.com}
}

\maketitle

\begin{abstract}

Recently, face recognition in the wild has achieved remarkable success and one key engine is the increasing size of training data. For example, the largest face dataset, WebFace42M contains about 2 million identities and 42 million faces. However, a massive number of faces raise the constraints in training time, computing resources, and memory cost. The current research on this problem mainly focuses on designing an efficient Fully-connected layer (FC) to reduce GPU memory consumption caused by a large number of identities. In this work, we relax these constraints by resolving the redundancy problem of the up-to-date face datasets caused by the greedily collecting operation (i.e. the core-set selection perspective).
As the first attempt in this perspective on the face recognition problem, we find that existing methods are limited in both performance and efficiency. For superior cost-efficiency, we contribute a novel filtering strategy dubbed Face-NMS. Face-NMS works on feature space and simultaneously considers the local and global sparsity in generating core sets. In practice, Face-NMS is analogous to Non-Maximum Suppression (NMS) in the object detection community. It ranks the faces by their potential contribution to the overall sparsity and filters out the superfluous face in the pairs with high similarity for local sparsity. With respect to the efficiency aspect, Face-NMS accelerates the whole pipeline by applying a smaller but sufficient proxy dataset in training the proxy model. As a result, with Face-NMS, we successfully scale down the WebFace42M dataset to 60\% while retaining its performance on the main benchmarks, offering a 40\% resource-saving and 1.64 times acceleration. The code is publicly available for reference at  \url{https://github.com/HuangJunJie2017/Face-NMS}.

\end{abstract}

\section{Introduction}

From MS1M (100K IDs, 10M faces)~\cite{MS1M} and Glint360K~\cite{Partial_FC}(360K IDs, 17M faces) to WebFace260M~\cite{WebFace}(4M IDs, 260M faces), face recognition datasets with increasing scale boost the performance of trained models, but at the same time, place a huge burden on storing and processing them. For example, training models on WebFace42M (2M IDs, 42M faces) would take 232G physical storage and 5,376 GPU hours~\cite{WebFace}. So, it has given rise to the study about how to reduce the training cost (the number of training machines, training time) caused by the large-scale identities and faces.

To tackle the dilemma, previous works mainly focus on designing an efficient FC layer~\cite{FaceNet,N_pair_Loss,Multi_Similarity_Loss,Circle_Loss}. Differently, we refer to the core-set selection technologies~\cite{coleman2020selection}, which come up with a novel perspective about data-efficient face recognition. Specifically, the greedy collection operation of the face data inevitably results in the presence of redundant data. By filtering out these redundant faces, we desire to explore a concise subset that performs as well as the entire set but is much more resource-saving than it. Although there are some core-set selection methods in image classification, applying them to the community of face recognition may lead to sub-optimal efficiency and performance degradation.

\begin{figure*}[t]
	\centering
    \subfloat[Random]{
    \label{fig:box}
    \begin{minipage}[c]{0.23\textwidth}
    \centering
    \centerline{\includegraphics[width=4cm,height=4cm]{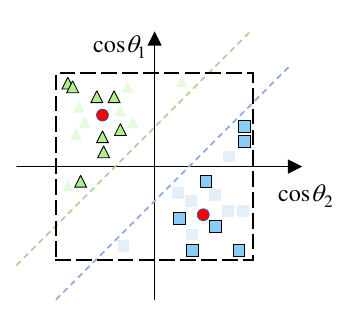}}
    \end{minipage}
    }
	\centering
    \subfloat[Sampling faces away from cluster center]{
    \label{fig:box}
    \begin{minipage}[c]{0.23\textwidth}
    \centering
    \centerline{\includegraphics[width=4cm,height=4cm]{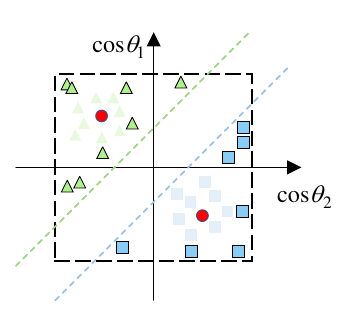}}
    \end{minipage}
    }
	\centering
    \subfloat[Removal of similar pairs]{
    \label{fig:box}
    \begin{minipage}[c]{0.23\textwidth}
    \centering
    \centerline{\includegraphics[width=4cm,height=4cm]{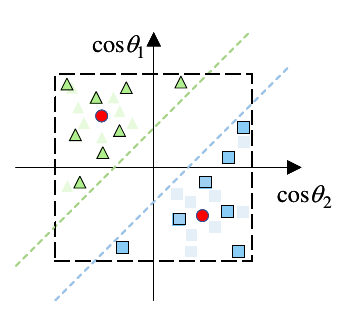}}
    \end{minipage}
    }
	\centering
    \subfloat[Face-NMS]{
    \label{fig:box}
    \begin{minipage}[c]{0.23\textwidth}
    \centering
    \centerline{\includegraphics[width=4cm,height=4cm]{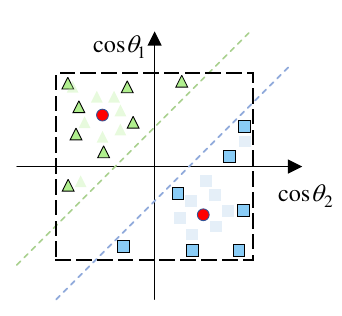}}
    \end{minipage}
    }
	\caption{Distribution of the core sets generated by different sampling strategies in the binary classification case. The dashed line represents the decision boundary. The red points represent the cluster center~\cite{AdaptiveFace} of each identity.}
	\label{fig:all_and_nms_samples}
\end{figure*}

Pursuing superior cost-efficiency, in this paper, we contribute a novel filtering strategy dubbed Face-NMS with an analogous mechanism as Non-Maximum Suppression (NMS)~\cite{NMS_1,NMS_2} in the object detection community. Face-NMS improves the core-set selection in the face recognition problem from both effectiveness and efficiency. Specifically, to resolve the redundancy problem, we pay attention to the face pairs with relatively high similarity in the feature space, where superfluous faces may exist. However, randomly dropping
within each similar face pair will lead to a sub-optimal result as it only guarantees the local sparsity. Pursuing a global sparsity, we reflect it by constructing an auxiliary indicator with all face pairs' cosine similarity. With this indicator, we theoretically prove that the faces away from the cluster center contribute more to the global sparsity. Accordingly, Face-NMS ranks the faces according to their similarity with the cluster center and tends to retain the faces away from the cluster center. In this way, by simultaneously considering local and global sparsity, Face-NMS can offer an optimal performance among the existing strategies when scaling down the datasets. As illustrated in Fig.~\ref{fig:all_and_nms_samples}, the core sets generated by Face-NMS are more reasonable with relatively high sparsity. With respect to the efficiency aspect, Face-NMS follows the selection-via-proxy paradigm proposed in \cite{coleman2020selection} with a small proxy model. However, the previous filtering strategies~\cite{ForgettingEvents,Activelearning,GreedyKCenters} require training the proxy model on the entire target dataset for some pivotal information. Rather than training on the target set, Face-NMS could train the proxy models on a proxy dataset that is far smaller than the target dataset as it works in the feature space. In this way, Face-NMS can offer an extra acceleration when compared with the previous works.

In the experiment, we apply Face-NMS to the ultra-large-scale WebFace42M~\cite{WebFace}. Though the dataset is scaled down to 60\%, the performance of the trained model is still comparable with the counterpart trained with the entire one. This means a 40\% saving of storage and computational resource. The storage requirement is scaled down from 232G to 139G and the computational resource requirement is scaled down from 5,376 GPU hours to 3,274 GPU hours. The main contributions can be summarized as follows:
\begin{itemize}

\item [1.]  We provide a new perspective to ease the burden of a large number of faces on training. To the best of our knowledge, this is the first attempt to study the mining of the core set in the community of face recognition.

\item [2.]  A novel filtering strategy dubbed Face-NMS is carefully designed with theoretical guidance. Compared with the pioneers, Face-NMS offers a higher cost-efficiency.

\item [3.] Comprehensive experiments are conducted to showcase the reasonableness and effectiveness of the proposed Face-NMS. On MS1M and ultra-large-scale WebFace42M, by sampling 60\% images per identity, our method attains almost the same performance with full data counterparts, reducing the training time and storage space to 60\%.

\end{itemize}
\section{Related Works}
\subsection{Efficient Face Recognition}
Deep neural networks have played an important role in the success of face recognition since DeepFace~\cite{DeepFace} employed convolutional neural networks on this task. There are two basic deep face recognition learning paradigms~\cite{Circle_Loss}. One is to classify the global identity of samples by applying a softmax loss~\cite{ArcFace,CosFace,SphereFace}. The other is to optimize the similarity between samples in batch by metric learning loss function~\cite{FaceNet,N_pair_Loss,Multi_Similarity_Loss,Circle_Loss}. As metric learning methods struggle in convergence and gain less in scaling up the dataset~\cite{Celeb500K}, softmax and its variants are the dominant paradigms in this field. However, when the number of identities grows, the size of the linear transformation matrix in the softmax-based method increases linearly, and the computational cost increases accordingly. As a result, softmax and its variants are suffering GPU memory limitation and computing resource consumption from the large-scale dataset.

To address the memory limitation, ArcFace~\cite{ArcFace} relieves the memory pressure on each GPU by applying model parallelism and computing the full-class softmax at the cost of little communication. Partial FC~\cite{Partial_FC} scales down the number of classes by randomly dropping negative classes in each mini-batch. As only the positive classes and the remaining negative classes are considered in the softmax calculation, the memory cost can be restricted in a feasible range which will not increase as the total class number increases. Recently, VFC~\cite{VFC} virtualizes FC parameters to reduce computational costs. DCQ~\cite{DCQ} employs a queue to dynamically select a subset of classes for each iteration during training.

\subsection{Core-set Selection}
Instead of resolving the memory limitation problem, we focus on the sample redundancy problem and study the core-set selecting strategy in this paper. The up-to-date famous datasets like MS1M~\cite{MS1M}, Celeb-500K~\cite{Celeb500K}, WebFace-260M~\cite{WebFace}, and so on are greedily collected via a list of celebrities and a search engine. The greedy collection aims at a complete result with as many samples as possible, where redundancy essentially exists. In the field of common classification problems, some works have been done to resolving the redundancy problem, which is also named core-set selection. Core-set selection can be broadly defined as techniques that find a subset of data points that maintain a similar level of quality (e.g., generalization error of a trained model or minimum enclosing ball) as the full dataset~\cite{coleman2020selection}. Some pioneering works indicate the effectiveness of core set selection. Namely, Soudry et al. \cite{boundary_1} argues that some data are more relevant to decision boundaries than others. Coleman et al. \cite{boundary_2} observes in experiments that models obtained by training a subset could reach or even exceed the full dataset.

The research community in core-set selection mainly focuses on the representative feature generation and the selecting strategy design. To obtain representative features for image classification, early works require ready-to-use features or training full target model itself~\cite{pioneer_1,pioneer_2,pioneer_3}. Recently, Coleman et al. \cite{coleman2020selection} deploys an efficient small proxy model to generate discriminative information like forgetting times~\cite{ForgettingEvents} and classification scores~\cite{Activelearning}. Yet, the proxy model is trained on the entire target dataset, which is resource consumptive but inevitable for these kinds of pivotal information. In this paper, we try to further improve the efficiency of this paradigm by designing a filtering strategy (i.e. Face-NMS) working on the feature space. In this way, we can train the proxy model on a proxy dataset that is far smaller than the target dataset. As there aren't any models needed to be trained on the full target dataset, the storage and computational requirement can be significantly reduced.

With respect to the core-set selecting strategy, it has been proven in \cite{boundary_1, boundary_2} that the core sets containing information-rich samples can perform well as the entire ones. Information used in this paradigm includes forgetting times~\cite{ForgettingEvents} and uncertainty~\cite{uty_1,uty_2}. These kinds of information work well in preserving global diversity but overlook the local redundancy problem. Besides, though Forgetting Events~\cite{ForgettingEvents} has been proved that can generate performance-degeneration-free core sets in \cite{coleman2020selection}, it performs poorly in scaling down the face set. This is because the prior of sufficient training epochs (e.g. performance stable after 75 epochs on CIFAR-10~\cite{cifar_10}) does not hold in face recognition where a relatively short schedule (e.g. 22 on MS1M ~\cite{ArcFace}) is adopted.
Some other methods~\cite{GreedyKCenters,diversity_1}, which attempt to approximate the distribution of the feature space of the entire dataset, have the potential to generate sparse core sets. By simulating the gradient of the full dataset, some other methods~\cite{Zhao2021DatasetCW, DD} synthesize a small informative image set. However, these methods are unfriendly to the large-scale dataset. Specifically, it is infeasible to perform Greedy K-Center~\cite{GreedyKCenters} on large-scale datasets like ImageNet~\cite{ImageNet} and MS1M~\cite{MS1M} due to its quadratic computational complexity~\cite{coleman2020selection}. Similarly, generating usable synthetic data requires thousands of loops, and these methods~\cite{Zhao2021DatasetCW,DD} have not been validated on large-scale datasets (e.g. ImageNet~\cite{ImageNet}) with large variations in appearance and pose.

Different from the existing methods, we resolve the core-set selection problem from the perspective of sparsity and seriously consider both local and global sparsity in designing our filtering strategy, which enables it to generate core sets that work as well as the whole set and outperform the existing methods. And our method presents effectiveness on large face sets~\cite{MS1M,WebFace}.

\section{Methodology}
We follow the selection-via-proxy paradigm proposed in \cite{coleman2020selection}. A proxy model is trained to extract the feature of samples for filtering. Then, the core set is generated by a filtering strategy and is used for training the target model. To select the core set from the dataset and maintain decent performance at the same time, we analyze the target of core-set selection to guide sampling faces in each identity at first. Then, based on the analysis result, we naturally formulate a novel sampling strategy called Face-NMS.

\subsection{The Target of Core-set Selection}
\label{Analyze}
Instead of searching for the most important sample, we resolve the core-set selection problem by dropping the redundant one from the perspective of sparsity. Given a face set of a specific identity $F = \{f_1,f_2,...,f_N\}$, where $f_i, i=1,...,N$ is the normalized feature vector of the $i^{th}$ face. The local redundancy exists in the pairs with high similarity as the faces with a close appearance contribute less to the overall diversity.
Thus, for local sparsity, it is intuitive to threshold the similarity between samples and randomly drop one face of the illegal pairs. However, the random dropping operation has not considered the differential of the contribution to the global sparsity between the two faces, which will lead to a sub-optimal result.

For comparison, the global sparsity can be defined according to the mean of the face pairs' cosine similarity within an identity.
\begin{equation}
\label{eq:diversity}
    S(F) = -\frac{1}{N^2}\sum_{i}^{N}\sum_{j}^{N}f_i\cdot f_j
\end{equation}
where $f_i\cdot f_j$ is the cosine similarity of the two face feature. Considering two extra samples $f$ and $f'$, we have:
\begin{equation}
\label{eq:contribute}
    \begin{split}
         &S(F \cup \{f\}) - S(F \cup \{f'\}) \\
        =& -\frac{2}{(N+1)^2}(\sum_i^N f_i\cdot f - \sum_i^N f_i\cdot f' )\\
        =& -\frac{2}{N+1}[f\cdot \frac{\sum_i^N f_i + f}{N+1}- f'\cdot \frac{\sum_i^N f_i + f'}{N+1}]
    \end{split}
\end{equation}
where $\frac{1}{N+1}(\sum_i^N f_i + f)$ is the feature mean of the faces in a specific identity (i.e. the cluster center). According to Eq.~\ref{eq:contribute}, the samples away from the cluster center will contribute more to the global sparsity. To maximize the global sparsity, instead of random choice, it is more reasonable to remove the samples close to the cluster center and retain the samples on the opposite. On the other hand, as illustrated in Fig.~\ref{fig:all_and_nms_samples}, the samples away from the cluster center form the boundary of the clusters, which is more important in inter-class learning for discriminant feature mining~\cite{NPCFace}. It is worth noting that selecting the samples according to their similarity with the cluster center can generate a core set with maximum scoring in Eq.~\ref{eq:diversity}. However, the local sparsity is ignored, which will also lead to a sub-optimal result.

In conclusion, the principle of core-set selection in face recognition is filtering out faces that share a high degree of similarity with some other and are close to the cluster center.

\begin{algorithm}[t]
	\caption {Face-NMS sampling}{$\mathcal{B}$ are faces that belong to an identity. $\mathcal{S}$ are similarity scores of faces to the cluster center. $N_t$ is the similarity threshold. $N$ is the number of faces in this identity.}

	\begin{algorithmic}[1]
		\STATE $\mathcal{B} = \{b_1, ..,b_N\} $, $\mathcal{S} = \{s_1, ..,s_N\}$
		\STATE $\mathcal{D} \leftarrow \{\}$
		\WHILE {$\mathcal{B} \ne$ empty}
		\STATE $m \leftarrow$ argmin $\mathcal{S} $
		\STATE $   b_r \leftarrow b_m,~~ \mathcal{B} \leftarrow \mathcal{B} - b_m$
		\STATE $ \mathcal{S} \leftarrow \mathcal{S} - s_m,~~ \mathcal{D} \leftarrow \mathcal{D} + b_m$
		  \FOR{$b_i ~ in ~ \mathcal{B}$}
		   \IF{~$similarity(b_r, b_i) \ge N_t$~}
		   \STATE $ \mathcal{B} \leftarrow \mathcal{B} -  b_i;~ \mathcal{S} \leftarrow \mathcal{S} -  s_i$\;
		   \ENDIF
		   \ENDFOR
		
		\ENDWHILE
		\STATE\textbf{return} {$\mathcal{D}$}

	\end{algorithmic}
    \label{alg:nms}
\end{algorithm}

\subsection{Face-NMS Sampling}\label{Face-NMS}

According to the aforementioned analysis, it is reminiscent of Non-Maximum Suppression (NMS) in detection~\cite{NMS_1,NMS_2}. NMS performs non-maximal result suppression to select detection results with relatively high confidence and filter out the redundant bounding boxes. Analogously, we propose a Face-NMS sampling strategy~ (Alg.~\ref{alg:nms}), which orderly selects faces according to similarity with the cluster center and drops faces with high similarity.

Specifically, the feature of each sample is extracted using the pre-trained proxy model, first. Then, the sampling strategy sorts the face list by the similarity score to the cluster center $\mathcal{S}$. After selecting the face $m$ with the minimum score, it is removed from set $\mathcal{B}$ and added to the final sampling set $\mathcal{D}$. Then, we calculate the similarity between face $m$ and other faces in set $\mathcal{B}$. Any face in set $\mathcal{B}$ with similarity to $m$ exceeding the threshold $N_t$ will be deleted. The above process is repeated until set $\mathcal{B}$ is empty. It is worth noting that since the proxy model works in feature space, it can be trained on a smaller proxy dataset rather than a redundant target dataset. As a result, the training process could be accelerated.

\section{Experiment}

\subsection{Experimental Settings} \label{settings}
\paragraph{Training Data.} We use two datasets MS-Celeb-1M (MS1M) and WebFace42M~\cite{WebFace}, as our training data. For MS1M, We apply the widely used MS1MV2~\cite{ArcFace}, a refined version of MS1M. The refined dataset contains 85K identities and 5.8M images. Its number of faces per identity is up to 68 and there are probably redundant samples. WebFace42M is a recently released large-scale face training data, containing 42M images, 2M identities. Since its number of identities is in million-scale, we only use it for final experiments due to the resource limitation. Additionally, we also evaluate the core sets' performance while the proxy models are trained on VGGFace2~\cite{VGGFace2} (3.3M images, 9K identities) or IMDB~\cite{IMDB} (1.0M images, 38K identities).

\paragraph{Testing Data.} We report the performance of the proposed method on the widely used benchmarks, including LFW~\cite{LFW}, CFP-FP~\cite{CFPFP}, AgeDB-30~\cite{AgeDB}, MegaFace~\cite{MegaFace}, and IJB-C~\cite{IJB_C}. MegaFace refers to rank-1 identification on \cite{MegaFace}. For IJB-C, we mainly focus on the performance of True Accept Rate at a False Accept Rate of 1e-4 (\textit{i.e.}, TAR@FAR=1e-4), which is considered as an objective metric with less affection from the noise~\cite{Xie2018ComparatorN}. And we also evaluate models trained on WebFace42M on the test set proposed in the same paper~\cite{WebFace} because it is less saturated. On this test set, we perform 1:1 verification and report the False Non-Match Rate (FNMR) under the False Match Rate (FMR) specified (lower FNMR at the same FMR is better)~\cite{WebFace}.

\begin{figure}[t]
		\centering
		\includegraphics[width=0.98\linewidth]{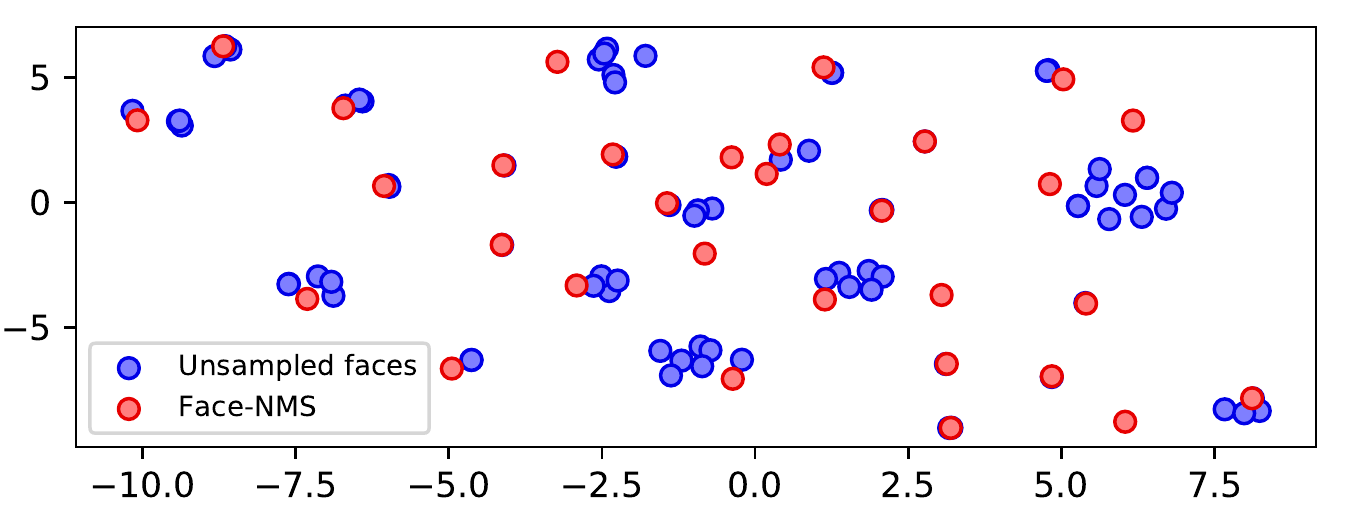}
		\caption{Visualization results of t-SNE~\cite{tSNE} for an identity in MS1M. Samples selected by Face-NMS are more dispersed than the unsampled faces (Best viewed in color).}
		\label{fig:tSNE}

\end{figure}

\paragraph{Implementation Details.}
The proxy model is constructed with ResNet-18~\cite{Resnet} backbone and is trained on MS1M~\cite{MS1M} by default. The target model is constructed with the ResNet-100 backbone and trained on MS1M or WebFace42M~\cite{WebFace}. Consistent with \cite{ArcFace} applied for face alignment, the five facial key-points predicted by Retinaface~\cite{Retinaface} are employed to generate normalized faces with a resolution of 112 $\times$ 112. CosFace (m=0.35)~\cite{CosFace} is used according to a united loss function~\cite{ArcFace}. During training, the learning rate is initially set to 0.1 and decreased by 10 at 10, 15, and 20 epochs. Optimizer SGD~\cite{SGD_1,SGD_2,SGD_3} is applied with a momentum of 0.9 and a weight decay of 5e-4. The batch size is set to 96 per GPU. The training process was terminated at 22 epochs. During training, we only adopt flip data augmentation. The similarity threshold ($N_t$ in Algorithm~\ref{alg:nms}) for selecting 60\% of the data is 0.80 on MS1M and 0.78 on WebFace42M.

\begin{figure}[t]
	\centering

	\includegraphics[width=0.46\textwidth]{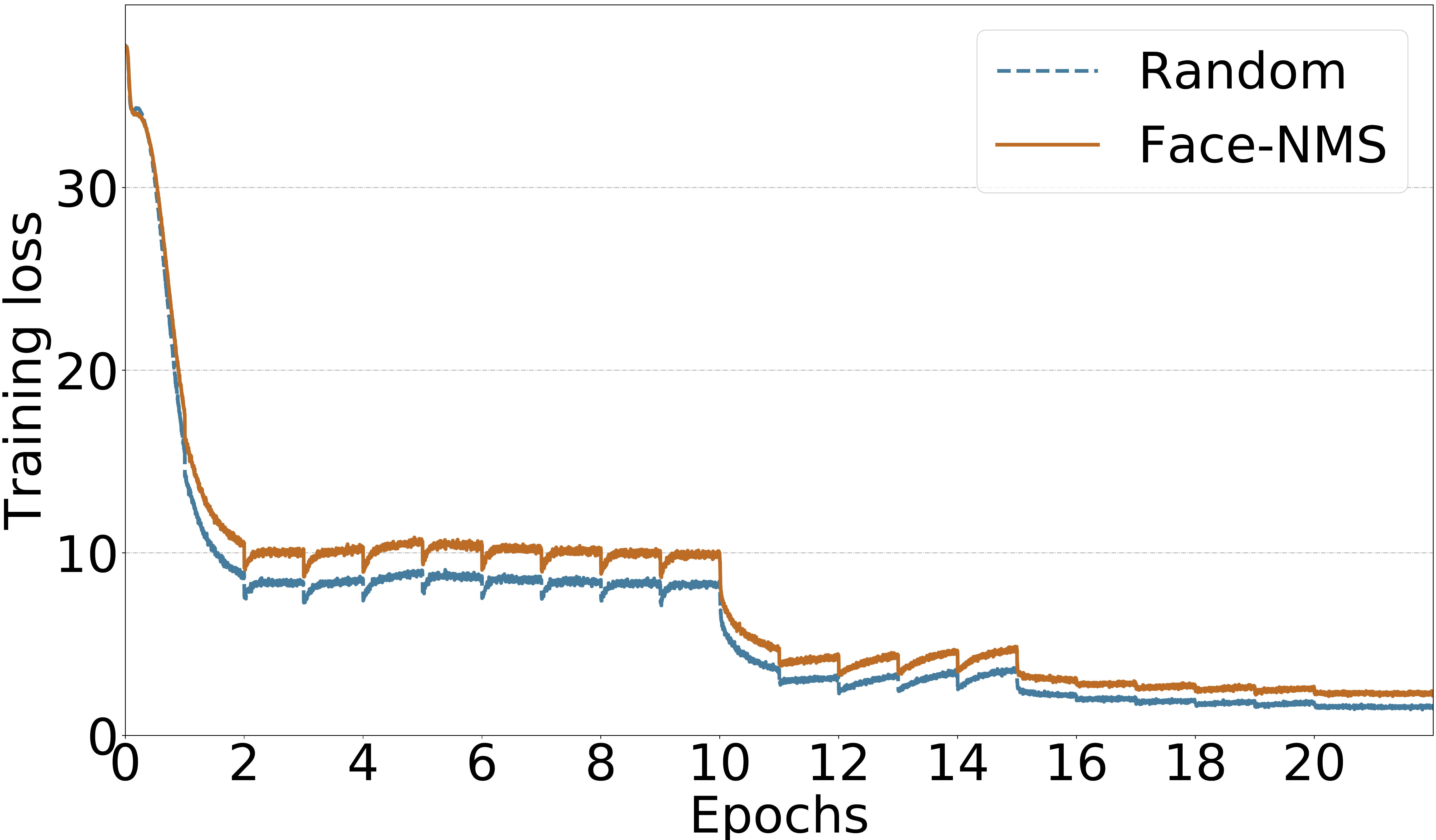}
	\caption{Training loss curves of Face-NMS sampling and random sampling applied on MS1M. Through Face-NMS, the training loss is consistently higher than random sampling.}	
		\label{fig:loss_curve}
\end{figure}

\subsection{Qualitative Evaluations}
\begin{figure}[t]
	\centering
	\includegraphics[width=\linewidth]{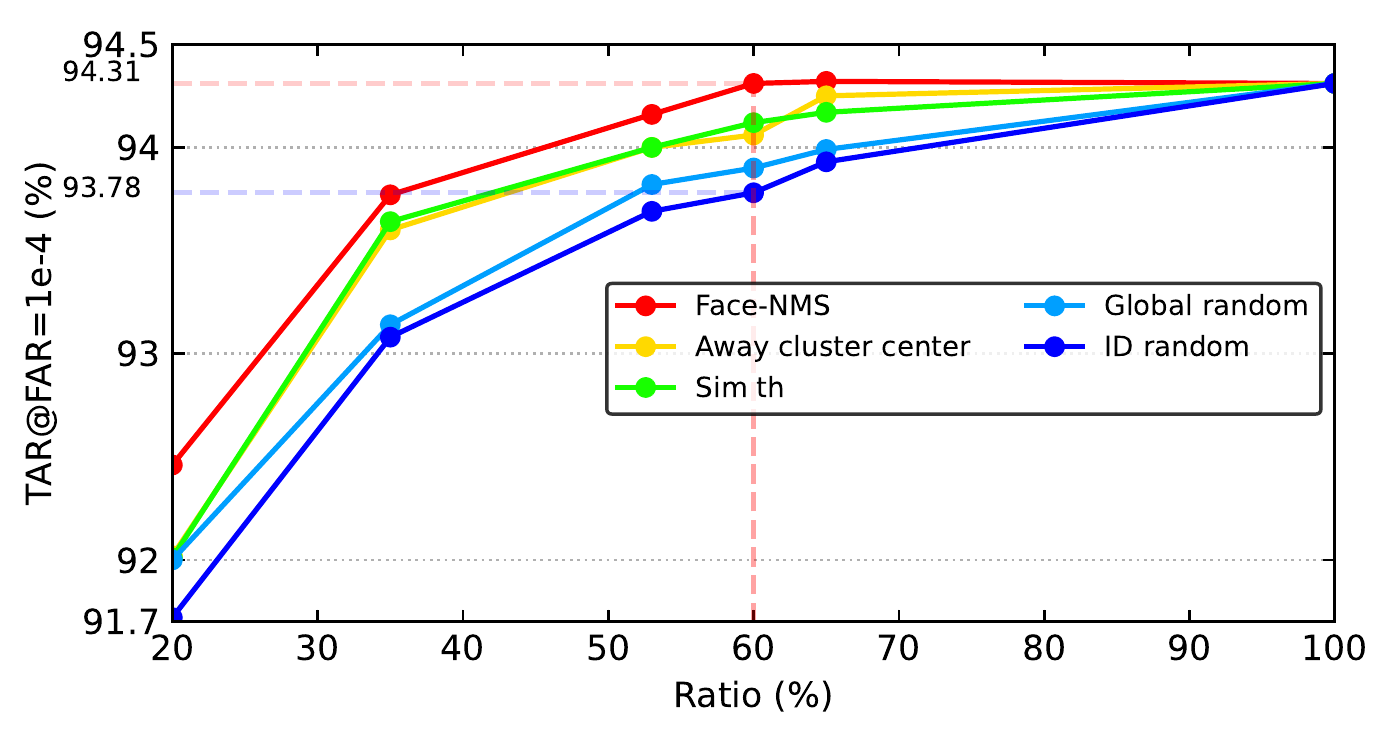}
	\caption{Performance of Face-NMS, Away cluster center sampling, Similarity threshold sampling (Sim th), Global random sampling, and Identity random sampling (ID random) under different sampling ratios. The horizontal axis represents the proportion of sampled faces to the full dataset. The vertical axis represents the performance on IJB-C (TAR@FAR=1e-4).}
	\label{fig:ablation_Face_NMS}
\end{figure}

We provide several perspectives in this subsection to showcase the problem we work on and the effectiveness of Face-NMS. The random sampling mentioned below corresponds to the global random sampling in Fig.~\ref{fig:ablation_Face_NMS}, i.e., randomly sample faces in the entire dataset. The number of faces from random sampling is equal to that of Face-NMS sampling.

\paragraph{The redundancy problem in face dataset}
In Fig.~\ref{fig:tSNE}, we visualize the sample distribution of a specific identity in feature space via t-SNE \cite{tSNE}. The corresponding face images are listed in supplementary materials for reference. According to the distribution of all samples (i.e. the blue dots), there are some clusters with high local density, which have high similarities with each other. They scale up the number of samples but contribute less to the overall diversity. By applying Face-NMS, the core set (i.e. the red dots) excludes most superfluous faces and uniformly distributes in the feature space, which is more reasonable.

\paragraph{Faces sampled by Face-NMS correspond to larger training loss values.}
In Fig.~\ref{fig:loss_curve}, we plot the training loss curves of different configurations on MS1M. Compared with the entire dataset, training on the core sets generated by Face-NMS has consistently higher training loss. Face-NMS tends to select faces away from the cluster center in feature space, which are prone to be hard samples in common sense~\cite{NPCFace,Mis_classified}.

\begin{figure}[t]
	\centering

	\includegraphics[width=0.46\textwidth]{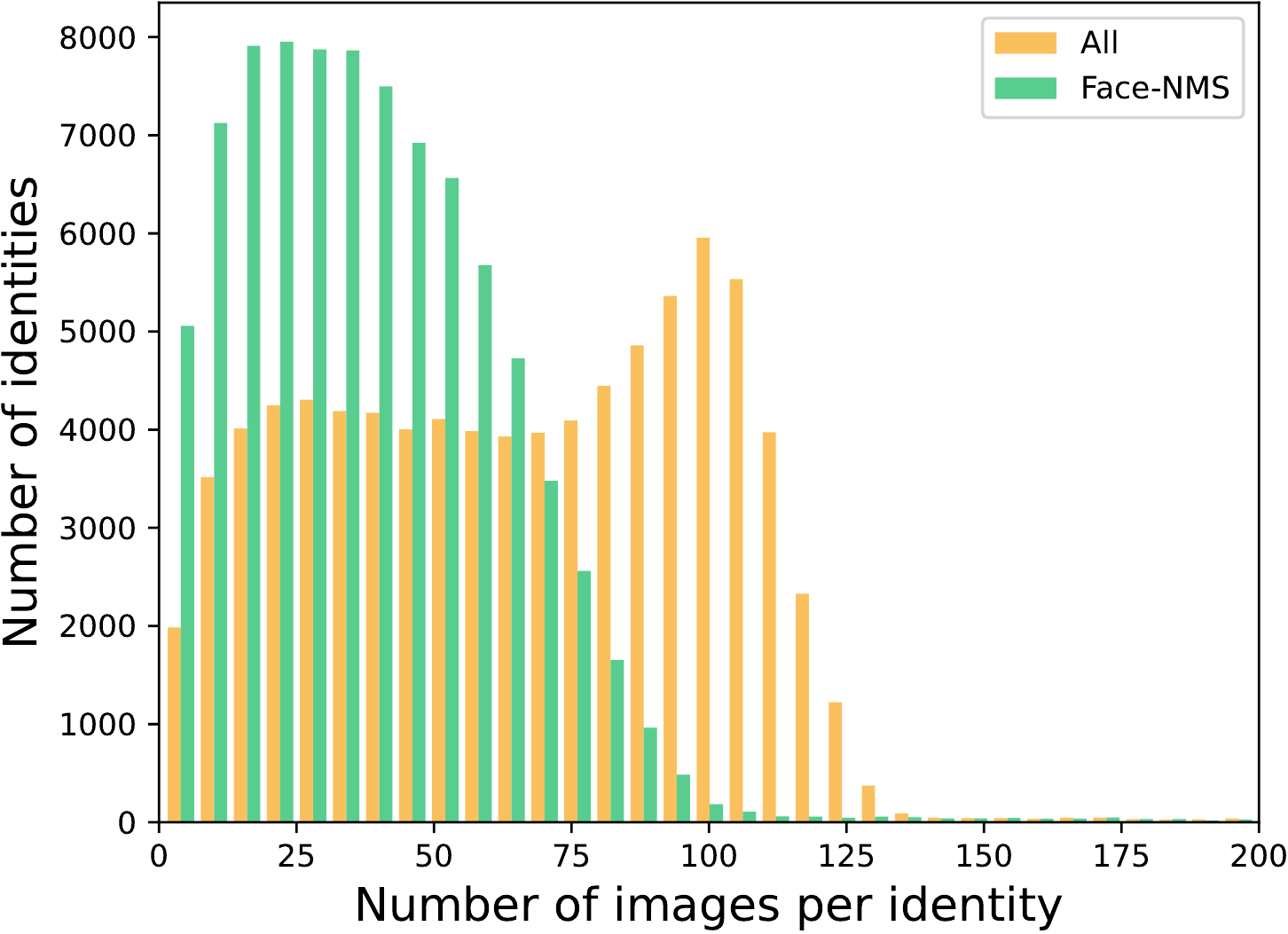}

		\caption{Distribution of the number of faces per identity before and after applying the Face-NMS sampling strategy on MS1M. Through Face-NMS, the mean and variance of the number of faces per identity on MS1M change from 67.91 $\pm$ 40.77 to 40.59 $\pm$ 30.40. With Face-NMS, the number of faces is more balanced among identities.}	
	\label{fig:num_dis}
\end{figure}

\paragraph{The number of faces per identity is more balanced.}
By removing redundant high-similarity faces by performing Face-NMS, the number of faces is more balanced among identities (Fig.~\ref{fig:num_dis}). Specifically on MS1M, by performing Face-NMS sampling, the variance of the number of faces per identity decreases from 40.77 to 30.40.

\begin{figure}[t]
	\centering
	\includegraphics[width=\linewidth]{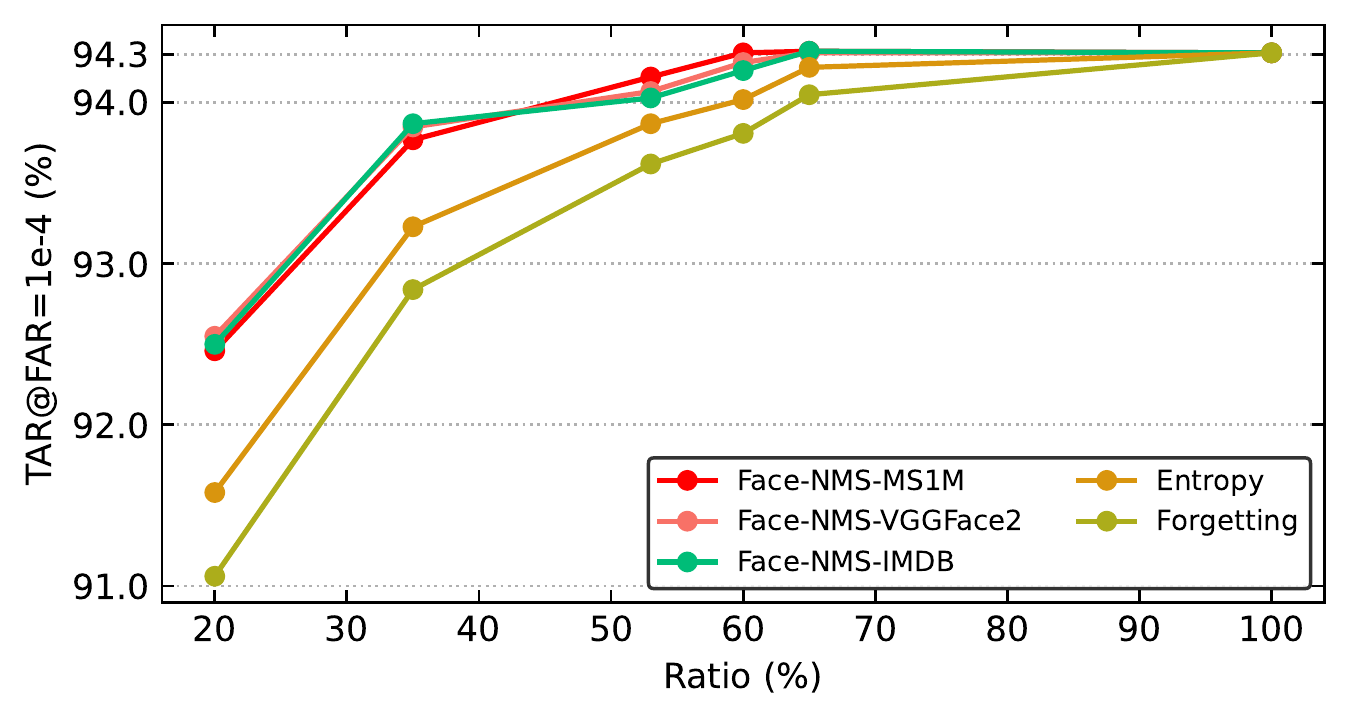}
	\caption{Performance of two core-set selection methods (entropy and forgetting) on MS1M. Besides, we list the performance of Face-NMS with proxy model trained on different datasets (MS1M, VGGface2 and IMDB).}
	\label{fig:small_model}
\end{figure}

\begin{table}[t]
  \centering
  \caption{Comparisons on MS1MV2 with state-of-the-art methods. We report the 1:1 verification
TAR (@FAR=1e-4) (\%) on the IJB-C dataset, rank-1 identification on MegaFace and the 1:1 verification accuracy (\%) on the LFW, CFP-FP, and AgeDB-30 datasets. ResNet-100 is used as backbone by default.}
  	\resizebox{\linewidth}{!}{
    \begin{tabular}{c|ccccc}
    \hline
    Model & IJB-C & MegaFace & LFW   & CFP-FP & AgeDB-30 \\
    \hline
    \hline
    CosFace(0.35)&  -     &  97.91     & 99.43  &     -  & - \\
    ArcFace(0.5) &  95.60 &  98.35     & 99.82  & 98.27  & - \\
    GroupFace      &    96.30   &  98.74     & 99.85  & 98.63  & 98.28 \\
    CurricularFace &    96.10   &  98.71     & 99.80  & 98.37  & 98.32  \\
    PartialFC-r1.0 &    96.40   &  98.36     & 99.83  & 98.51  & 98.03  \\
    PartialFC-r0.1 &    96.00   &  98.31    & 99.82  & 98.60  & 98.13  \\
    \hline
    SVP-Entropy (60\%) & 95.83  & 97.51  & 99.77  & 98.13  & 97.88  \\
    SVP-Forgetting (60\%) & 95.76  & 97.09  & 99.68  & 98.04 & 97.53   \\
    \hline
    CosFace (Ours,100\%) &  96.22     &   \bf98.20    &  99.73     &  98.20     & \bf97.95  \\
    Random (Ours,60\%) &  95.69     & 97.74      & 99.73      &  97.94     & 97.50 \\
    Face-NMS (Ours,60\%) &  \bf96.23     &  98.15     &  \bf99.80     &  \bf98.40     & 97.93 \\
    \hline
    \end{tabular}%
    }
  \label{tab:MS1M_efficient}%
\end{table}%

\begin{table}
  \centering
  \caption{Comparison of models trained with WebFace42M dataset. Test set of \cite{WebFace} is evaluated under FNMR@FMR=1e-5. And lower FNMR at the same FMR is better. ResNet-100 is used as backbone by default.}
    	\resizebox{1.0\linewidth}{!}{
    \begin{tabular}{c|ccccc|c}
    \hline
    WebFace42M & IJB-C & MegaFace & LFW   & CFP-FP & AgeDB-30 & Test Set \\
    \hline
    \hline
    All       &   \bf97.83    &   99.02       &  \bf99.83     &  \bf99.39      &  \bf98.53 & \bf0.0256\\
    Random (60\%)   &   97.78     &  98.99     &  \bf99.83     &  99.34      &  98.45 & 0.0273\\ 
    Face-NMS(60\%) &    97.82    &  \bf99.03     &  \bf99.83     &  \bf99.39      &  98.47 & \bf0.0256\\
    \hline
    \end{tabular}%
    }
  \label{tab:WebFace42M}%
\end{table}%

\begin{table*}[t]
  \centering
  \caption{Computational complexity (CC) and storage complexity (SC) required in training the target model by directly on WebFace42M (Baseline), SVP~\cite{coleman2020selection} and Face-NMS.}
  \resizebox{\linewidth}{!}{
    \scriptsize
    \begin{tabular}{c|ccc|ccc|c|c|c}
    \hline
    \multirow{2}{*}{Method} & \multicolumn{3}{c|}{Proxy Model} & \multicolumn{3}{c|}{Target Model} & \multicolumn{1}{c|}{\multirow{2}{*}{Total CC}} & \multirow{2}{*}{Speed-up} & \multirow{2}{*}{SC} \\
\cline{2-7}          & Backbone & Training Data  & CC  & Backbone &  Training Data  & CC  &       &       &  \\
    \hline
    \hline
    Baseline & -     & -     & -     & R100  & WebFace42M(100\%) & 5,376 GPU hours   & 5,376 GPU hours  & 1.00x  & 232G \\
    SVP   & R18   & WebFace42M(100\%) & 1,604 GPU hours & R100  & WebFace42M(60\%) & \bf3,226 GPU hours  & 4,830 GPU hours  & 1.11x  & 232G \\
    Ours  & R18   & MS1M(100\%) & \bf48 GPU hours   & R100  & WebFace42M(60\%) & \bf3,226 GPU hours  & \bf3,274 GPU hours  & \bf1.64x  & \bf139G \\
    \hline
    \end{tabular}%
    }
  \label{tab:datasetstatistics}%
\end{table*}%

\subsection{Benchmark Results}


We apply the Face-NMS sampling strategy on two training datasets, MS1MV2 and WebFace42M. Trained models are evaluated on LFW~\cite{LFW}, CFP-FP~\cite{CFPFP}, AgeDB-30~\cite{AgeDB}, MegaFace~\cite{MegaFace}, and IJB-C~\cite{IJB_C}.

\paragraph{Comparisons on MS1MV2.}
We compare the mainstream methods of both efficient face recognition and core-set selection in Tab.~\ref{tab:MS1M_efficient}. Following most SOTA methods, we choose ResNet-100 as the training backbone. As illustrated in Tab.~\ref{tab:MS1M_efficient}, by sampling 60\% of faces, Face-NMS achieves almost the same performance as the full dataset on various test sets which is comparable with the existing efficient face recognition methods. Face-NMS significantly outperforms random sampling on all testing sets. For example, the TAR (@FAR=1e-4) of Face-NMS reaches 96.23\% on IJB-C, while the performance of random sampling is 95.69\%.

For the existing methods of core-set selection, we try to deploy the three selecting strategies introduced in \cite{coleman2020selection}. However, since the greedy K-center approach simultaneously requires quadratic computational complexity and quadratic storage complexity~\cite{coleman2020selection}, it is prohibitively expensive to perform it on large-scale datasets with millions of samples like ImageNet~\cite{ImageNet} and MS1M~\cite{MS1M}. Thus, we merely employ the other two methods dubbed SVP-Entropy and SVP-Forgetting on the face recognition task. The performance comparison of different core-set selecting strategies is illustrated in Fig.~\ref{fig:small_model}. At all testing scales, both SVP-Entropy and SVP-Forgetting fail to maintain the performance of the target models. They also have sub-optimal performance on the others test set as listed in Tab.~\ref{tab:MS1M_efficient}. For example, with the same configuration of sampling 60\% faces, SVP-Entropy and SVP-Forgetting score 95.83\% and 95.76\% on IJB-C, which is excelled by Face-NMS (96.23\%). With respect to the face recognition problem, Face-NMS has superior performance among existing core-set selecting methods.

\paragraph{Comparisons on WebFace42M.}
For the models trained on WebFace42M, it is not convincing to evaluate them on the conventional test sets due to the limitations of nearly saturated performance. As listed in Tab.~\ref{tab:WebFace42M}, although 60\% data sampled by Face-NMS achieves almost the same performance as the full dataset, the difference between 60\% randomly sampled data and the whole set is marginal. For example, the discrepancy between the two is only 0.05\% on IJB-C (from 97.78\% to 97.83\%) and 0.03\% on MegaFace (from 98.99\% to 99.02\%). Thus, we apply the testing protocol proposed in the paper~\cite{WebFace} to further evaluate the models trained on WebFace42M. On this test set, the model trained with 60\% images of WebFace42M generated by Face-NMS still attains almost the same performance with full data. It exceeds the performance of random sampling by 0.17\% FNMR@FMR=1e-5 (lower FNMR at the same FMR is better).

\paragraph{Reduction in dataset storage space and training time.}
In Tab.~\ref{tab:datasetstatistics}, we take the model with ResNet-100 backbone and trained it on WebFace42M as the target. The computational complexity and the storage complexity of the baseline configuration are 5376 GPU hours and 232G, respectively. SVP~\cite{coleman2020selection} takes the model with ResNet-18 backbone and trained on WebFace42M as the proxy models, and trained the target models on the core set with 60\% faces of the whole set. This configuration takes 1604 GPU hours for training the proxy models and 3226 GPU hours for training the target models. A total of 4830 GPU hours merely has an acceleration of 1.11 times on the baseline. As the computational complexity of the final fully-connected layer is proportional to identity number, it takes a large proportion in the overall computational complexity when training models on WebFace42M. Thus, replacing the ResNet-100 backbone with the ResNet-18 backbone just offers a small acceleration on the whole pipeline. Besides, as the proxy model is trained on the whole WebFace42M, the storage complexity of SVP is still 232G. Without performance degeneration, Face-NMS further accelerates this paradigm by applying a proxy dataset (i.e. MS1M) in training the proxy model, which reduces the computational complexity of training the proxy model from 1604 GPU hours to merely 48 GPU hours. The overall computational complexity can be reduced to 3274 GPU hours, offering an acceleration of 1.64 times on the baseline. Besides, as no longer needed to train any model on the whole WebFace42M, the storage complexity is reduced from 232G to 139G.

\begin{figure}[t]
	\centering
	\includegraphics[width=\linewidth]{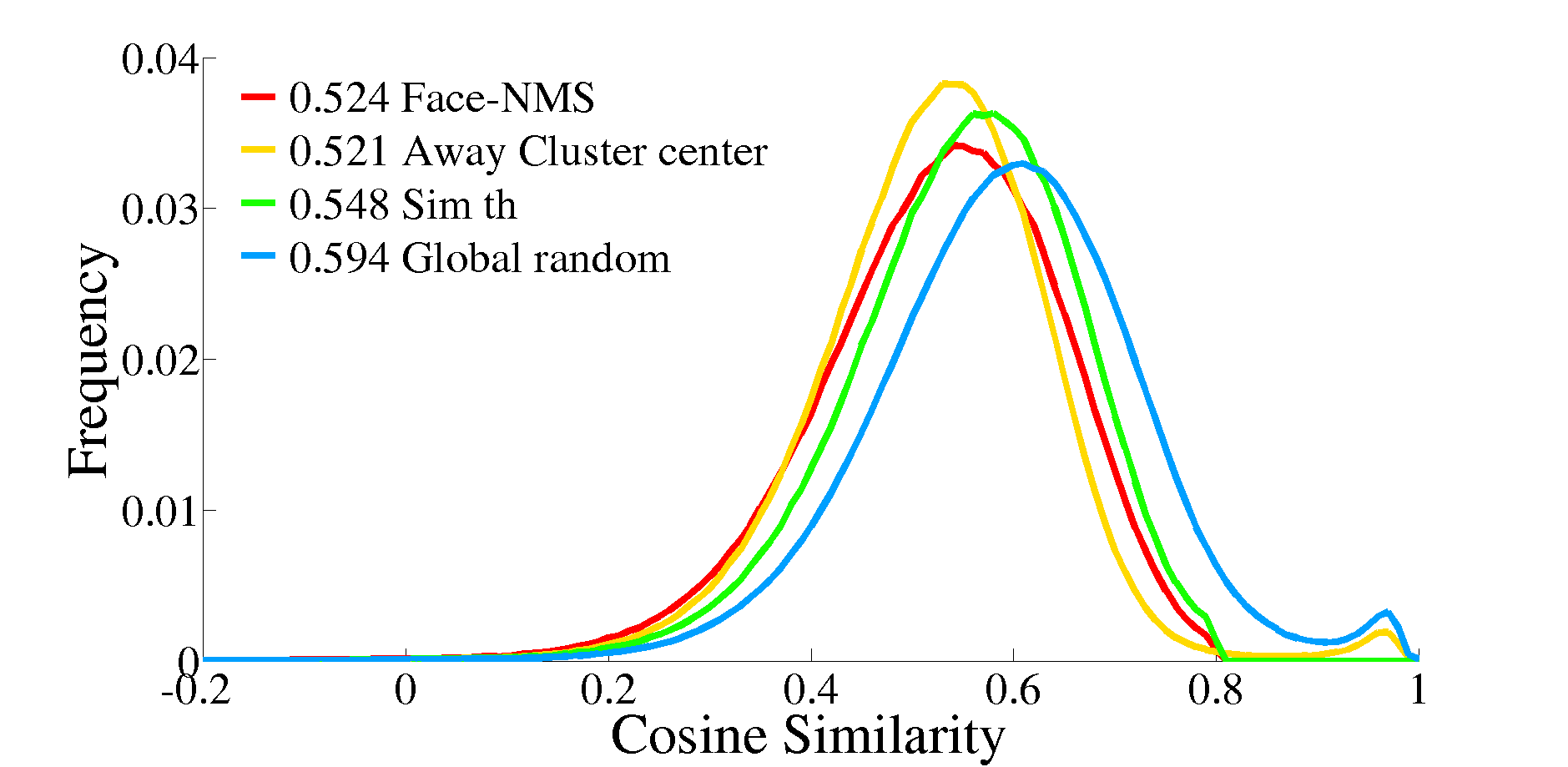}
	\caption{Intra-class cosine similarity distribution of the core sets generated by Face-NMS, away cluster center sampling, similarity threshold sampling (Sim th), and global random sampling. The vertical axis is the normalized frequency. The mean of each curve is listed in the corresponding legend. }
	\label{fig:Face_NMS_effect}
\end{figure}

\subsection{Ablation Studies}
\paragraph{The necessity of considering both global and local sparsity.}
To showcase the necessity of considering both global and local sparsity of the core set, we construct five sets of controlled experiments for comparison. The core sets generated by part of these methods are illustrated in Fig.~\ref{fig:all_and_nms_samples}. The corresponding intra-class similarity distribution is partially illustrated in Fig.~\ref{fig:Face_NMS_effect}. (1) Face-NMS sampling considers both local and global sparsity in filtering faces. (2) Away cluster center sampling: ignore the local sparsity and select faces far away from the cluster center. The core sets generated by this configuration have the highest global sparsity. However, some similar face pairs exist and degrade the local sparsity. (3) Similarity threshold sampling (Sim th): threshold the feature similarity of faces pairs in each identity and randomly drop one face in each illegal pair. Only local sparsity is considered and the global sparsity is ignored in this configuration. The core set generated by this configuration has a relatively low global sparsity when compared with that generated by Face-NMS. (4) Global random sampling: put all faces together and randomly sample a proportion from it. Both local and global sparsity are not considered in this configuration. (5) Identity random sampling (ID random): randomly sample a proportion of faces in each identity. Both local and global sparsity are not considered in this configuration.

The performances of the corresponding target models under different core-set scales are illustrated in Fig.~\ref{fig:ablation_Face_NMS} for comparison. When sampling over 60\% faces, the core-set generated by Face-NMS works well as the whole set and consistently outperforms other configurations. Core-set generated by similarity threshold sampling and away cluster center sampling both have sub-optimal performance in all testing scales. This indicates that both local and global sparsity should be seriously considered when scaling down the dataset to a small scale. Ignoring either local or global sparsity will lead to performance degradation. When both of them are not considered in generating core-set, scaling down the dataset by randomly sampling faces in each identity or in the whole dataset has worse performance.

\paragraph{The affect of the proxy datasets.}
It has been proven in \cite{coleman2020selection} that acceleration can be achieved by applying a proxy model with a small backbone. This also holds in the face recognition problem. Besides, as the proposed method Face-NMS works in feature space, we can gain extra acceleration by applying a small proxy dataset for training the proxy model. In this subsection, we study the effect of the proxy datasets by applying IMDB~\cite{IMDB} and VGGface2~\cite{VGGFace2} with a scale of 17\% MS1M and 57\% MS1M, respectively. The performance of different proxy datasets is illustrated in Fig.~\ref{fig:small_model}. According to the experimental results, scaling down the proxy dataset has little impact on the core-set performance at all testing scales. Thus, Face-NMS is robust to the performance of the proxy models and has the potential to gain extra acceleration by applying a small proxy dataset for training the proxy model. We just verify this potential in this paper. The lower bound on the size of the proxy dataset has not been studied and will be done in future work. In addition, Due to space limitations, the performance of the proxy model and the time required to train it are presented in the supplemental material.

\section{Conclusion}
In this work, we achieve data-efficient face recognition by shooting light on the redundancy problem in the up-to-date training sets. By formulating an effective filtering strategy Face-NMS, which equips the core set with both local and global sparsity, we successfully scale down the training set by 40\%, while maintaining its performance on popular benchmarks. Besides, through training the proxy model on the proxy set instead of the target set, Face-NMS reduces the storage requirement and preserves the computational resource. Future works will focus on (1) Exploiting a more effective core-set selecting strategy for face recognition. (2) Exploring the lower bound on the scale of the proxy dataset. (3) Resolving the inter-class redundancy problem.

{\small
\bibliographystyle{ieee_fullname}
\bibliography{aaai22}
}
\end{document}